\documentclass[letterpaper,10pt,conference]{ieeeconf} % IEEE format two column
\IEEEoverridecommandlockouts
\usepackage{float}
\usepackage{graphicx}
\usepackage{amssymb}

\usepackage{amsmath}
\usepackage{mathtools}  
\usepackage{algorithmic}
\usepackage{algorithm}
\usepackage{nccmath}

\usepackage{amsthm}
\usepackage[usenames,dvipsnames,svgnames,table]{xcolor}
\theoremstyle{plain}

\newtheorem{definition}{Definition}
\newtheorem{problem}{Problem}

\usepackage{float}
\floatstyle{plaintop}
\restylefloat{table}
\usepackage{MnSymbol}%
\usepackage{wasysym}%
\usepackage{xspace} 
\usepackage[hidelinks]{hyperref}

\newcommand{\new}[1]{{\color{black}#1}}

\hypersetup{
    colorlinks=false,
    linkcolor=black,
    filecolor=black,      
    urlcolor=black,
    }

\begin{document}

\title{ \LARGE \bf Non-Prehensile Manipulation of Cuboid Objects Using \\a Catenary Robot
}

\author{ Gustavo A. Cardona$^*$, Diego S. D'Antonio$^*$, Cristian-Ioan Vasile, and David Salda\~{n}a 
\thanks{${}^*$ The authors contributed equally.}
\thanks{Gustavo A. Cardona, Diego S. D'Antonio, Cristian-Ioan Vasile and David Salda\~{n}a are with the Autonomous and Intelligent Robotics Laboratory -AIRLab- at Lehigh University, PA, USA:$\{$\texttt{gcardona, diego.s.dantonio, cvasile, saldana\}@lehigh.edu}}        

}

\maketitle
\thispagestyle{empty}
\pagestyle{empty}

\begin{abstract}
Transporting objects using quadrotors with cables has been widely studied in the literature. However, most of those approaches assume that the cables are previously attached to the load by human intervention.
In tasks where multiple objects need to be moved, the efficiency of the robotic system is constrained by the requirement of manual labor.
Our approach uses a non-stretchable cable connected to two quadrotors, which we call the catenary robot, that fully automates the transportation task.
Using the cable, we can roll and drag the cuboid object (box) on planar surfaces. Depending on the surface type, we choose the proper action, dragging for low friction, and rolling for high friction. Therefore, the transportation process does not require any human intervention as we use the cable to interact with the box without requiring fastening.
We validate our control design in simulation and with actual robots, where we show them rolling and dragging boxes to track desired trajectories.

\end{abstract}

\IEEEpeerreviewmaketitle

%-------------------------------------------------------------
\section{Introduction}
%-------------------------------------------------------------

Nowadays, aerial transportation and manipulation using quadrotors have become topics of great interest due to the low cost and efficiency in applications such as search and rescue~\cite{cacace2016control}, drone delivery~\cite{bamburry2015drones}, commercial cargo, home assistance, and precision agriculture \cite{villa2019survey}.
High maneuverability, high thrust-to-weight ratio, improvements in on-board computing power, and battery efficiency have made quadrotors advantageous for performing manipulation tasks.
 
Approaches that tackle transportation and manipulation using quadrotors include~\cite{kim2013aerial,fanni2017new, ollero2021past}.
The two major strategies in the literature are either equipping the quadrotor with a gripper or fastening device~\cite{suarez2017anthropomorphic}, or using cables attached to the load~\cite{sreenath2013geometric}.
On the one hand, quadrotors connected to loads via rigid connections have simpler models and controllers for object manipulation. Nevertheless, they are more susceptible to disturbances and uncertainties due to the direct interaction between bodies \cite{mendoza2020snake,zhao2018design}.
On the other hand, cable-suspended approaches consider the load either as a point-mass concentration~\cite{cardona2019cooperative} where all of the attaching cables go to the same point or a rigid body with multiple attaching points~\cite{li2021cooperative}.
The latter adds degrees of freedom in the load and avoids swinging, but it increases the complexity of the dynamics and control.

\begin{figure}[t]
\centering
\includegraphics[width = 0.45\textwidth]{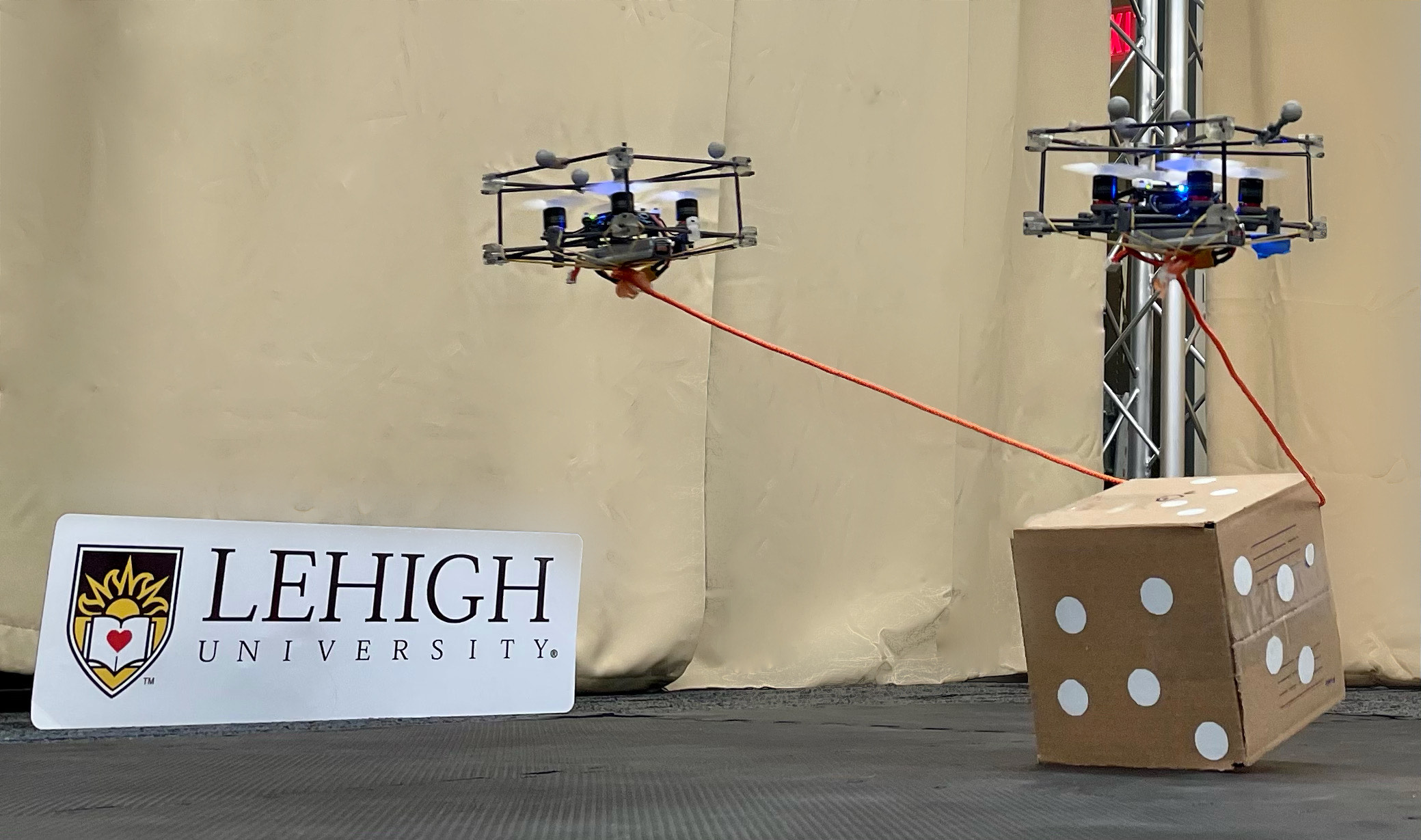}
\caption{A catenary robot rolling and dragging a box. \newline Video:~\url{https://youtu.be/Ou6DPlXPE7A}}
\label{fig:real}
\end{figure}

Cable-suspended approaches, in particular, consider cables to be previously attached to objects, which is not always feasible or desirable when deploying quadrotors in real environments. Avoiding human intervention to attach objects requires identifying the way to establish contact with the object. Becoming crucial to know the better location where the forces should be applied~\cite{donald2000distributed}, since factors like the slipping between objects and coulomb friction forces~\cite{trinkle1990planning} might affect the manipulation maneuver. In contrast to having fastening devices or previously attached cables, we propose to use a flexible cable whose ends are attached to quadrotors that steer the cable to form contacts with objects. We refer to the cable and the quadrotors as the \textit{catenary robot}. We use the cable as a non-prehensile manipulator, since it is not used to  grasp or hold the object.

Flexible cables propelled by quadrotors attached at their ends are considered in the literature, modeled as a chain where every link affects the subsequent links and it is appended as a new body in the system \cite{kotaru2018differential, goodarzi2015geometric}.  However, this model requires high computational resources due to the increased number of variables in the system's dynamics. In this work, we propose using the natural shape of a flexible cable when it hangs from its ends due to gravity, known as a catenary. This shape allows us to control the catenary robot with the less computational effort since we can determine the cable's parameters, its span and lowest point by knowing the quadrotors' location.
Once the cable creates tension with the object, we model the cable as straight lines instead of the caternary curve.

We develop non-prehensile manipulation algorithms to move a box on a planar surface using the catenary robot, as shown in Fig.~\ref{fig:real}.
We consider two manipulation modes, dragging and rolling. Dragging is well suited when friction between the box and the surface is low.
In this case, the cable is placed lower than the box's center of mass height to ensure continuous contact with the surface and smooth motion.
On the other hand, rolling is advantageous when the friction is higher. 
Thus, the cable is placed close to the higher edge maximizing the torque over the box, rolling the box over an edge into a new position.

% \david{I would remove the following paragraph. The catenary robot is not a contribution of this paper. Experiments are not contributions.
% The new contribution is the new way to interact with objects without requiring fastening, but it is described earlier in the intro.}
% The main contributions of the paper are threefold.
% First, we propose a versatile robot called the catenary robot composed of a cable attached to two quadrotors whose shape, defined by span and lowest point, can be
% estimated and controlled from the quadrotors' positions.
% Second, we propose two control strategies to manipulate boxes on planar surfaces via dragging and rolling.
% Finally, we show the performance of our algorithms in simulated and experimental trials.

%------------------------------------------------------------
%\section{Preliminaries and Assumptions}
\section{Problem Statement}
\label{Sec: Problem}
%-------------------------------------------------------------
The goal of this work is to perform non-prehensile manipulation using a pair of quadrotors and a cable.
We use a catenary robot tasked with moving cuboid objects on a planar surface.

%\subsection{The Catenary Robot}

\begin{definition}[\textbf{Catenary Robot}]
\label{def:flying-rope}
A Catenary Robot is a pair of quadrotors connected by a non-stretchable cable of length $\ell$ with its ends attached to the center of mass of each quadrotor.
\end{definition}

\begin{figure}[t]
\centering
\includegraphics[width = 0.35\textwidth]{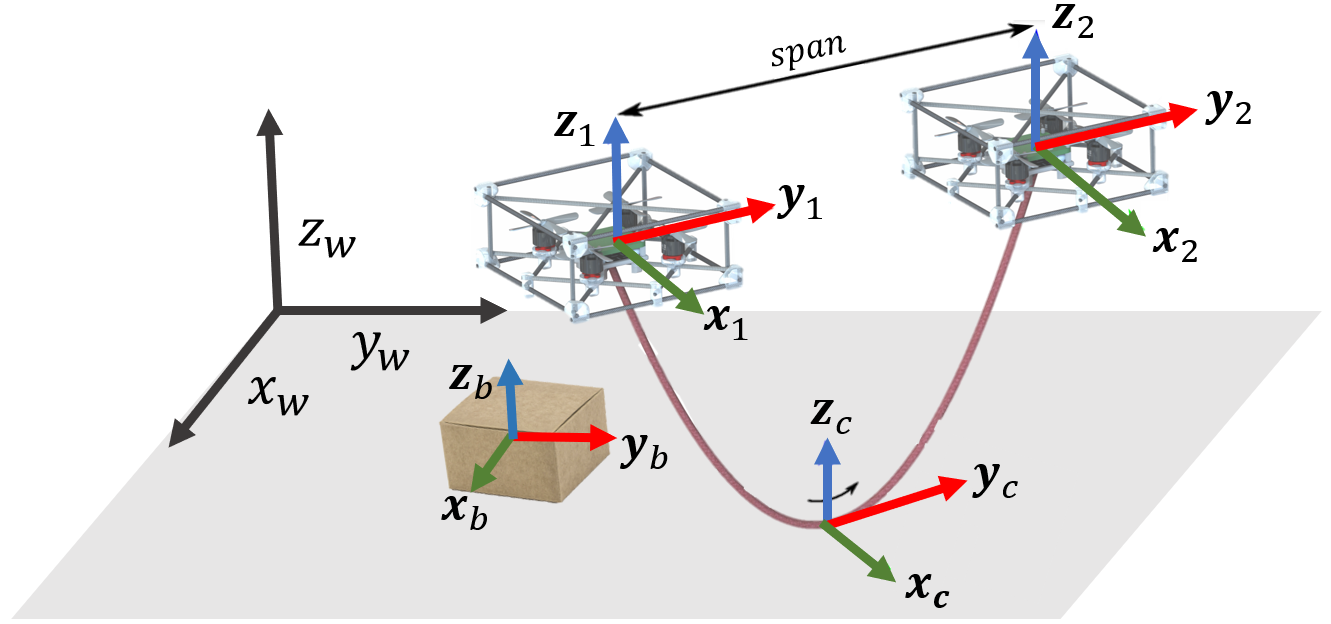}
\caption{Catenary robot, a box on a flat surface and their coordinate frames.}	
\label{fig:problem}
\end{figure}

The quadrotors are identical, same actuators, mass $m$, and inertia matrix $\mathbf{J}$.
The name \textit{catenary robot} comes from the name of the curve that the cable describes when it is held by its ends.
We illustrate the catenary robot, a box, and the coordinate frames in Fig. \ref{fig:problem}. The coordinate frames are: 
world frame or inertial frame, with the $z$-axis pointing upwards, denoted by $\{w\}=\left \{ \mathbf{x}_w, \mathbf{y}_w, \mathbf{z}_w \right \}$;
quadrotor frame, for $i=1,2$, located at the quadrotor's center of mass with the $x$-axis pointing to the front and the $z$-axis pointing upwards, denoted by $\{q_i\}=\left \{ \mathbf{x}_{i}, \mathbf{y}_{i}, \mathbf{z}_{i} \right \}$;
box frame, located at the box's center of mass with the $z$-axis pointing to the upper face, and the $x$-axis pointing to the front face in the direction of motion,
denoted by $\{b\}=\left \{ \mathbf{x}_b, \mathbf{y}_b, \mathbf{z}_b \right \}$; 

For each quadrotor $i\in \{1,2\}$,
we denote its location in the world frame by $\mathbf{r}_{i} \in \mathbb{R}^3$, s.t. $\|\mathbf{r}_{1}-\mathbf{r}_{2}\|\leq \ell$, where $\ell$ is the length of the cable,
and its angular velocity in $\{q_i\}$  by $\boldsymbol{\omega}_i \in \mathbb{R}^{3}$. 
The force experienced by the $i$th quadrotor  due to the cable's tension is $\mathbf{t}_i$ in $\{w\}$.
\new{Each quadrotor} can produce  thrust
$f_i \in \mathbb{R}$ and torque  $\boldsymbol{\tau}_i\in \mathbb{R}^{3}$. 
Therefore, the control input of the catenary robot is
$(f_1, f_2, \boldsymbol{\tau}_1, \boldsymbol{\tau}_2)$.
The dynamics of each quadrotor  is described by the Newton-Euler equations
\begin{align}
%X_{free}\colon=\left \{  
%\begin{aligned}
m {\mathbf{\ddot{r}}}_{i} &=-m g \mathbf{e}_3+{f}_i {}^{w}\mathbf{R}_i\mathbf{e}_3+ {}^{w}\mathbf{R}_{i} \mathbf{t}_i,\\ 
\mathbf{J}\dot{\boldsymbol{\omega}}_i &=\mathbf{J}{\boldsymbol{\omega}}_i \times \boldsymbol{\omega}_i + \boldsymbol{\tau}_i\new{,}
%\end{aligned}\right .
\label{eq2:dynamicquad}
\end{align}
where $^{a}\mathbf{R}_b$ is the rotation matrix in the frame $a$ of the body $b$.
% As described previously our goal in is to move a box on a planar surface. 
% The box is affected by gravity, friction, and the tension generated by interaction with the cable.
The catenary robot uses its cable to interact with cuboid objects.

\begin{definition}[Box]
A \emph{box}  is a cuboid object with 
width, length\new{,} and height; $w,l,h\in \mathbb{R}_{>0}$ respectively. 
\end{definition}
It has a mass $m_b \in \mathbb{R}_{>0}$, and inertia tensor $\mathbf{J}_b \in \mathbb{R}^{3 \times 3}$. Its position 
%, velocity, and acceleration  of the box 
in  $\{ w\}$ is $\mathbf{r}_b \in \mathbb{R}^{3}$
%, $\mathbf{\dot{r}}_b \in \mathbb{R}^{3}$, and $\mathbf{\ddot{r}}_b \in \mathbb{R}^3$  
and its angular velocity in $\{ b\}$ is $\boldsymbol{\omega}_b$.
The catenary robot can use the cable to pull the box in different directions.
When the cable touches the vertical edges of the box, there is at least one intersecting point between the cable and a vertical  edge. We denote the closest intersecting point to the $i$th quadrotor by $\mathbf{p}_i$ in $\{b\}$.
%The tension of the cable  produced by the box to the $i-{th}$ quadrotor is denoted by $\mathbf{f}_{T_i} \in \mathbb{R}_{\geq 0}$.
The friction force between the box and the ground is $\mathbf{f}_b$. 
Since the tension in the quadrotor frame is $\mathbf{t}_{i}$, the box feels the opposite tension $-\mathbf{t}_{i}$.
Thus, the dynamics of the box is described by
\begin{align}
  m_{b} {\mathbf{\ddot{r}}}_{b} &=-m_{b} g \mathbf{e}_3-\sum_{i}{}^{{w}}\mathbf{R}_{\new{b}} \mathbf{t}_{i}  +{}^{{w}}\mathbf{R}_{\new{b}}\mathbf{f}_b,\\ 
\mathbf{J}_b\dot{\boldsymbol{\omega} }_{b}&=
\mathbf{J}_b{\boldsymbol{\omega}}_b \times \boldsymbol{\omega}_b -\sum_{i}\mathbf{t}_{i} \times \mathbf{p}_{i}\new{.}
\end{align}

\subsection{Actions}
A cable is highly versatile, offering multiple ways to interact with objects, e.g. pulling, tightening, dragging. In this paper, we focus on non-prehensile manipulation \cite{ruggiero2018nonprehensile} by performing two main actions: dragging and rolling.
We employ the catenary robot to perform these actions and move the box on a planar surface.

\begin{definition}[Dragging action]
\emph{Dragging} is the action of translating or rotating the box on the ground. It involves motion on the $xy$-plane and rotation with respect to the $z$-axis of $\{w\}$.
\end{definition}
Note that here the box is not able to roll along the $y$-axis or the $x$-axis.  
Dragging the box is advantageous when the friction force is low, e.g., slippery surface like ice, since the pulling forces from the robot directly contribute to the box's motion. 

\begin{definition}[Rolling action]
\emph{Rolling} is the action of translating the box by making it rotate around its $x$- and $y$-axis.
\end{definition}
This action is relevant when the friction force with the ground is high, making it difficult to drag. 
In this paper, we want to use the catenary robot to manipulate a box on a planar surface autonomously. This leads us to the following problem.

\begin{problem}
Given a box with dimensions $(w,l,h)$, mass $m_b$ on a planar surface,
use the control input of the catenary robot $(f_1, f_2, \boldsymbol{\tau}_1, \boldsymbol{\tau}_2)$ to perform rolling and dragging actions.
\end{problem}  

We consider changes on the ground material that might affect the friction force experienced by the box, forcing the catenary robot to switch between \textit{dragging} or \textit{rolling}.
However, if the friction constraints are met, it is possible to change from dragging to rolling
just by manipulating the contact points.

%-------------------------------------------------------------
\section{Interaction Between the Robot and the Box}
\begin{figure}[b]
\centering
\includegraphics[width = 0.99\linewidth]{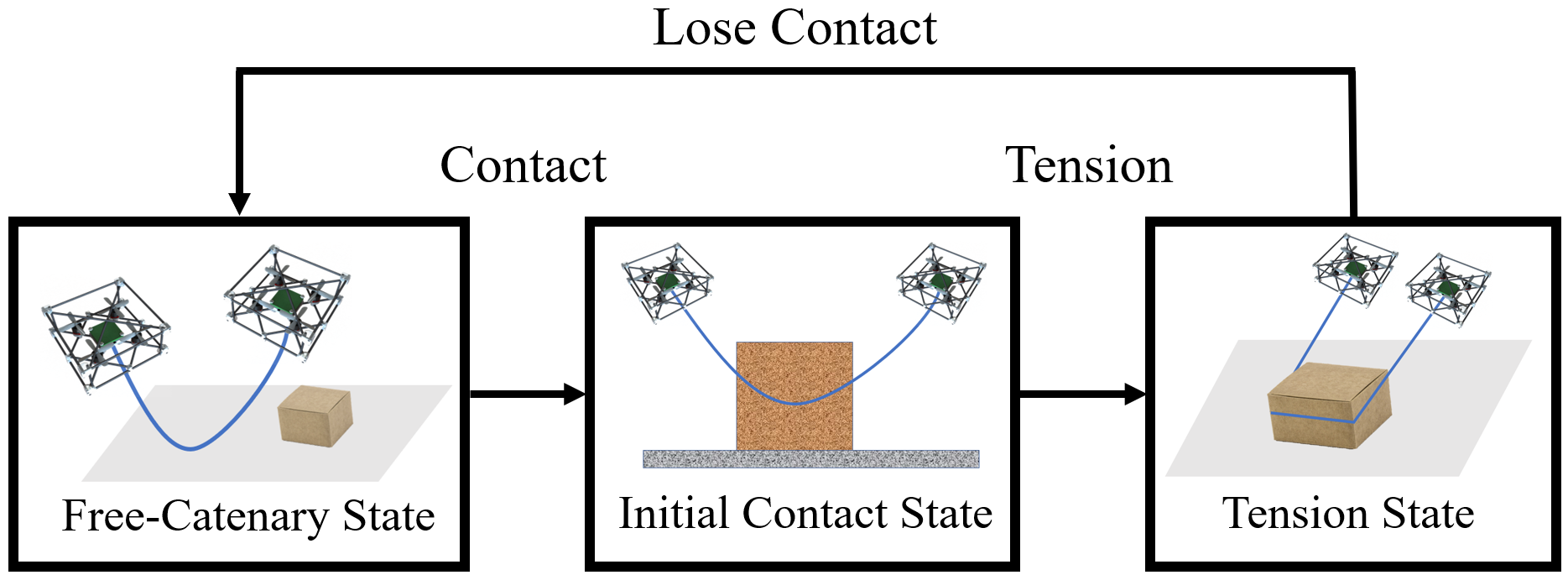}
\caption{States based on the contact between the cable and the box.}	
\label{fig:states}
\end{figure}

In order to interact with the box, we define three main states (see Fig.~\ref{fig:states}):
\subsubsection{Free-catenary state} In this state, the cable is not in contact with the box or the ground.
The objective of this state is to move the robot near to the object.
The catenary robot can move freely in the environment and is not having any interaction with the box.
Thus, the configuration space \new{of each body (two quadrotors and one box)} is $SE(3)$.
In our previous work~\cite{catenaryrobot}, we presented a controller to track
a desired trajectory for the catenary robot in the free-catenary state, specified by its reference point $\mathbf{c}(t)$, orientation $\psi(t)$, and span $s(t)$.
Hence, we can move the robot to approach the box.
This paper focuses on the following states when there is contact between the cable and the box. 

This strategy allows to place the cable on the desired contact point.
Depending on whether the box is going to be dragged or rolled, the altitude changes.
By changing the distance between the quadrotors the tension on the hanging cable changes.
Getting higher tensions in the cable offers a better opportunity to avoids slipping the cable on the box, making it easier to place the cable at the desired contact point.
Once the cable touches the box, we switch to the next state.

\subsubsection{Initial-contact state} 
\begin{figure}[t]
    \centering
    \includegraphics[width = 0.33\textwidth]{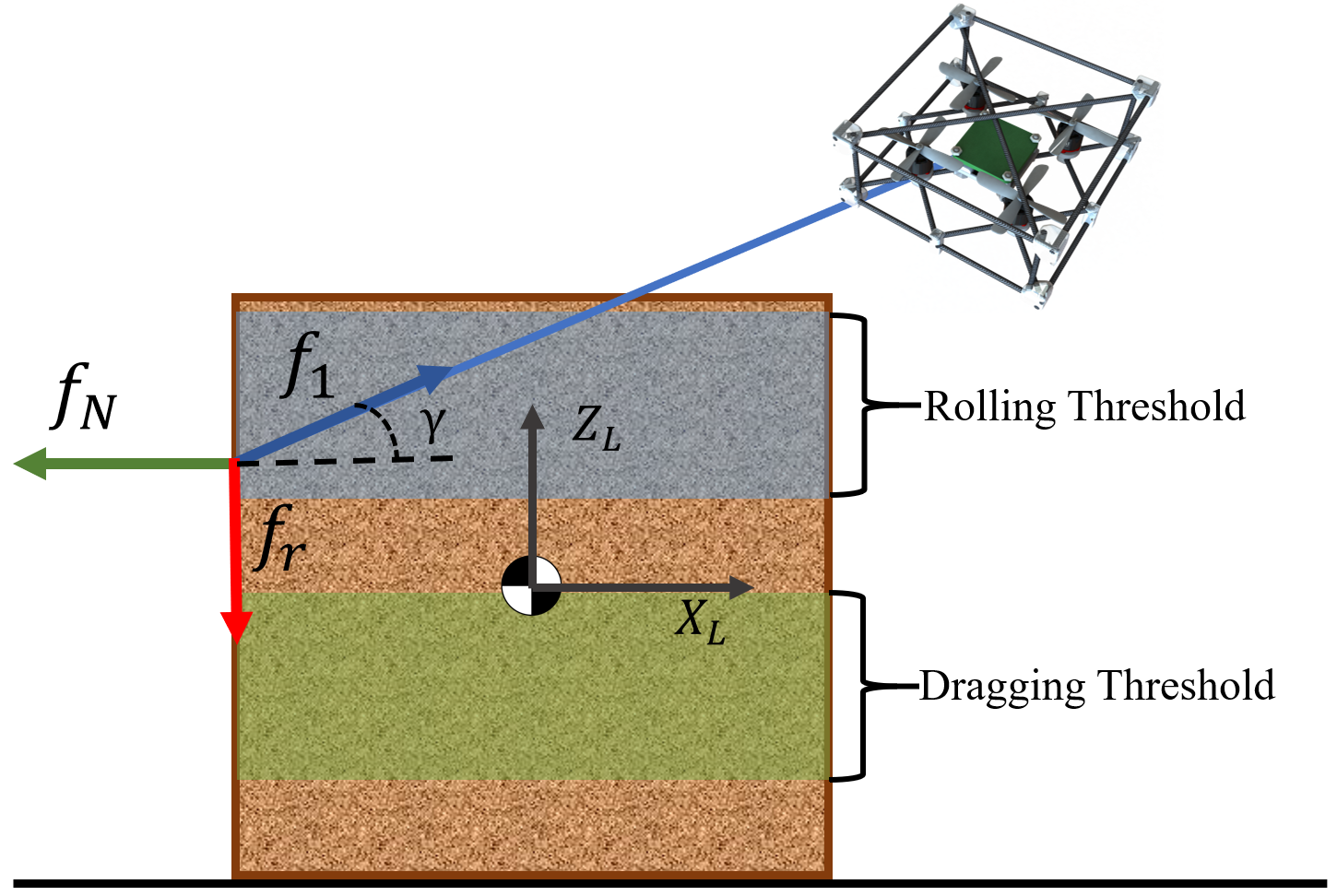}
    \caption{Contact model: the contact interaction between the cable and the box depends on the direction of the forces and the friction experienced by the two materials. We consider the blue area as ideal to perform rolling, and the green area to perform dragging.}
    \label{fig2:contact}
\end{figure}

This state starts when the robot is in free-state, approaches the box and then the cable point $\mathbf{c}(t)$ touches the vertical face of the box. The location of the point $\mathbf{c}(t)$ on the box frame and the span $s(t)$ determines the effect of the tension on the box.
Therefore, depending on the action, either dragging or rolling, we need the optimal way to approach the box.
It is also worth taking into account that the point of contact between the box and the cable affects directly the amount of force needed to move the box. In Fig. \ref{fig2:contact}, we define the dragging threshold where the cable should be located in order to maximize the linear movement of the box. The angle $\gamma$ of this point of contact to the quadrotors is also important and is discussed later.
Once the cable changes from the catenary shape to form three straight lines due to the tension, we switch to the next state.

\subsubsection{Action state} In this state, the cable is in contact with the box, and the cable between the box and the quadrotors is completely straight.
Friction between the cable and the box is critical since depending on the tension vector the cable can slip loosing the contact with the box.

\begin{figure*}[t]
\centering
$
\begin{array}{ccc}
     \includegraphics[width = 0.33\textwidth]{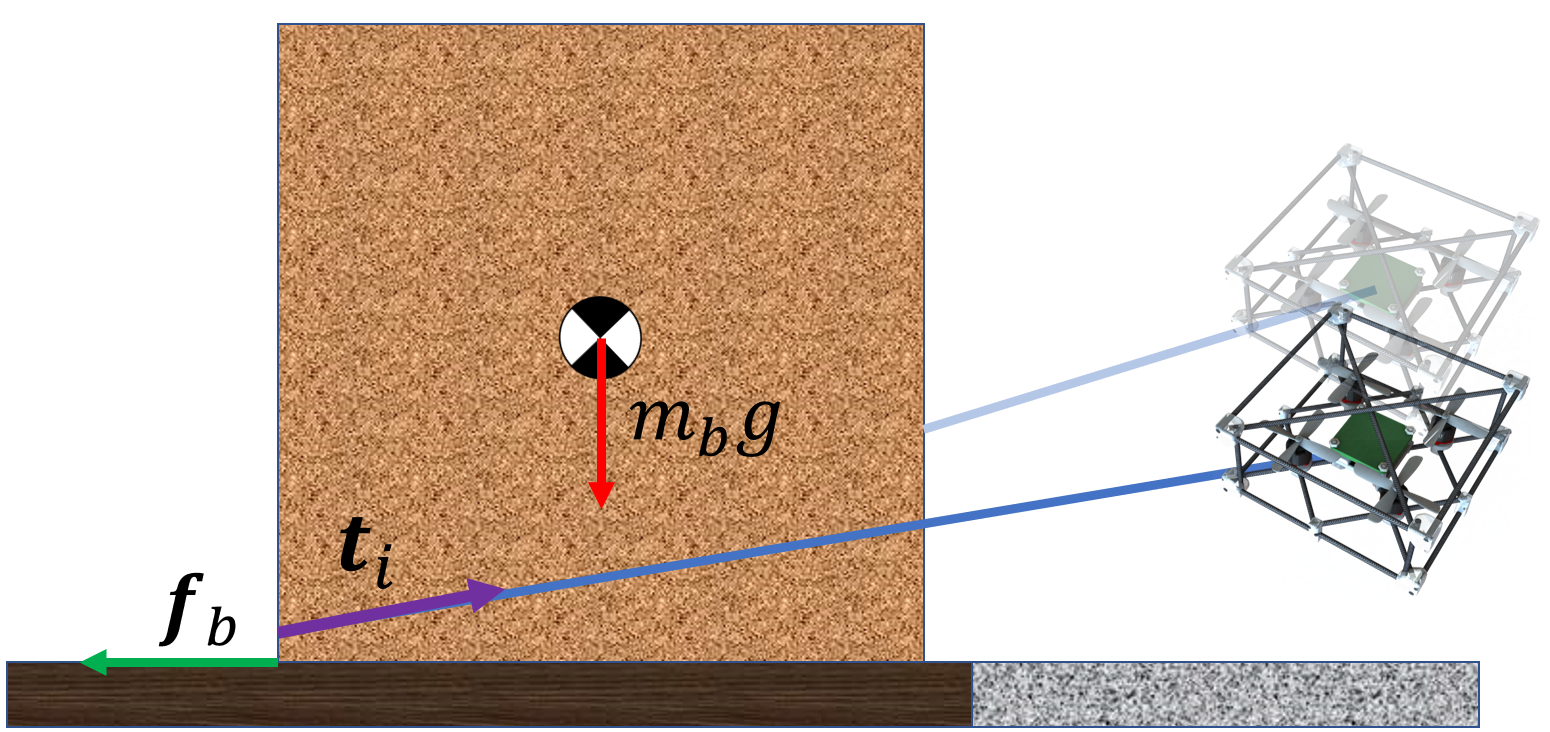} \label{subfig:drag1}
     & \includegraphics[width = 0.33\textwidth]{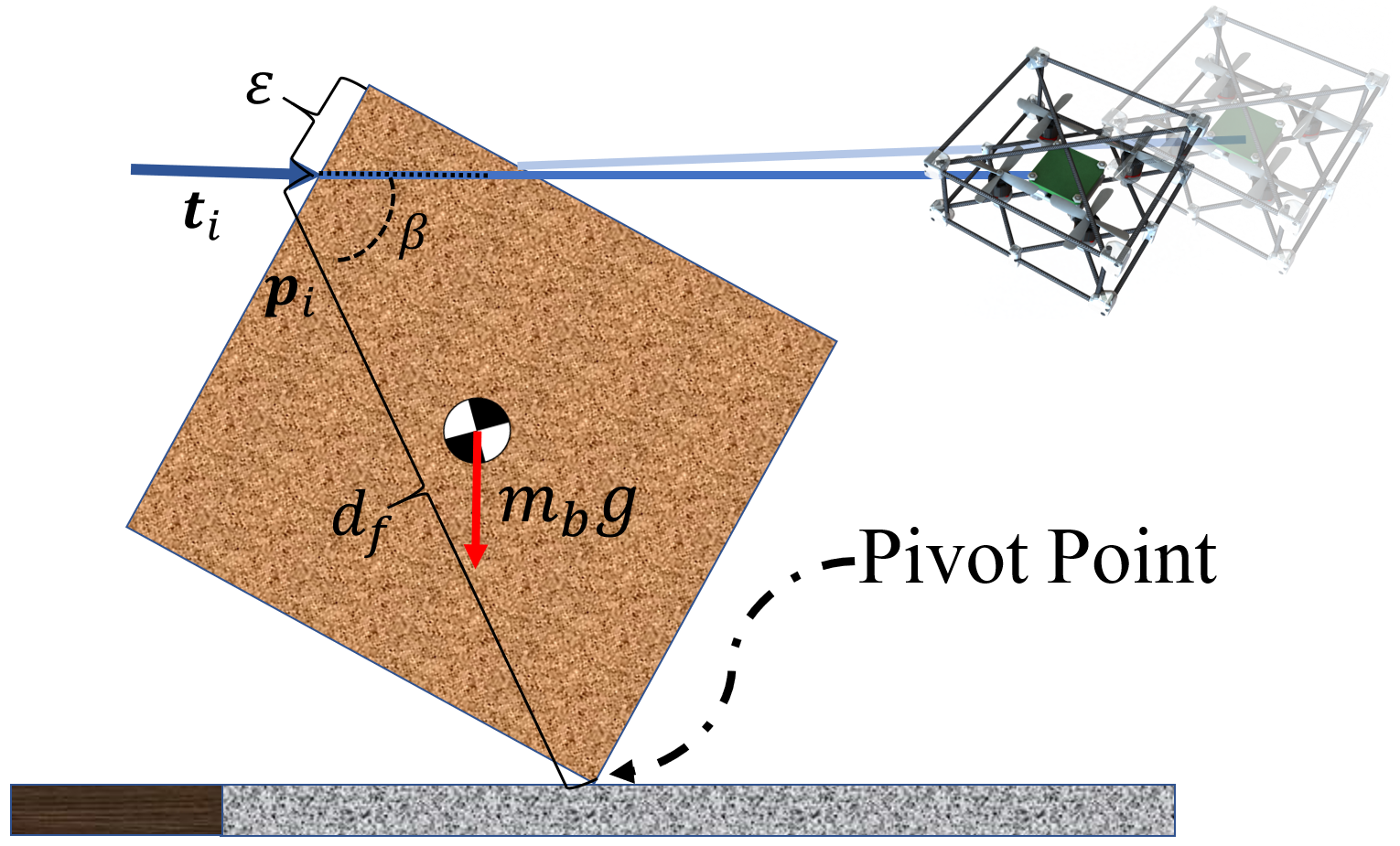}
     & \includegraphics[height = 0.2\textwidth]{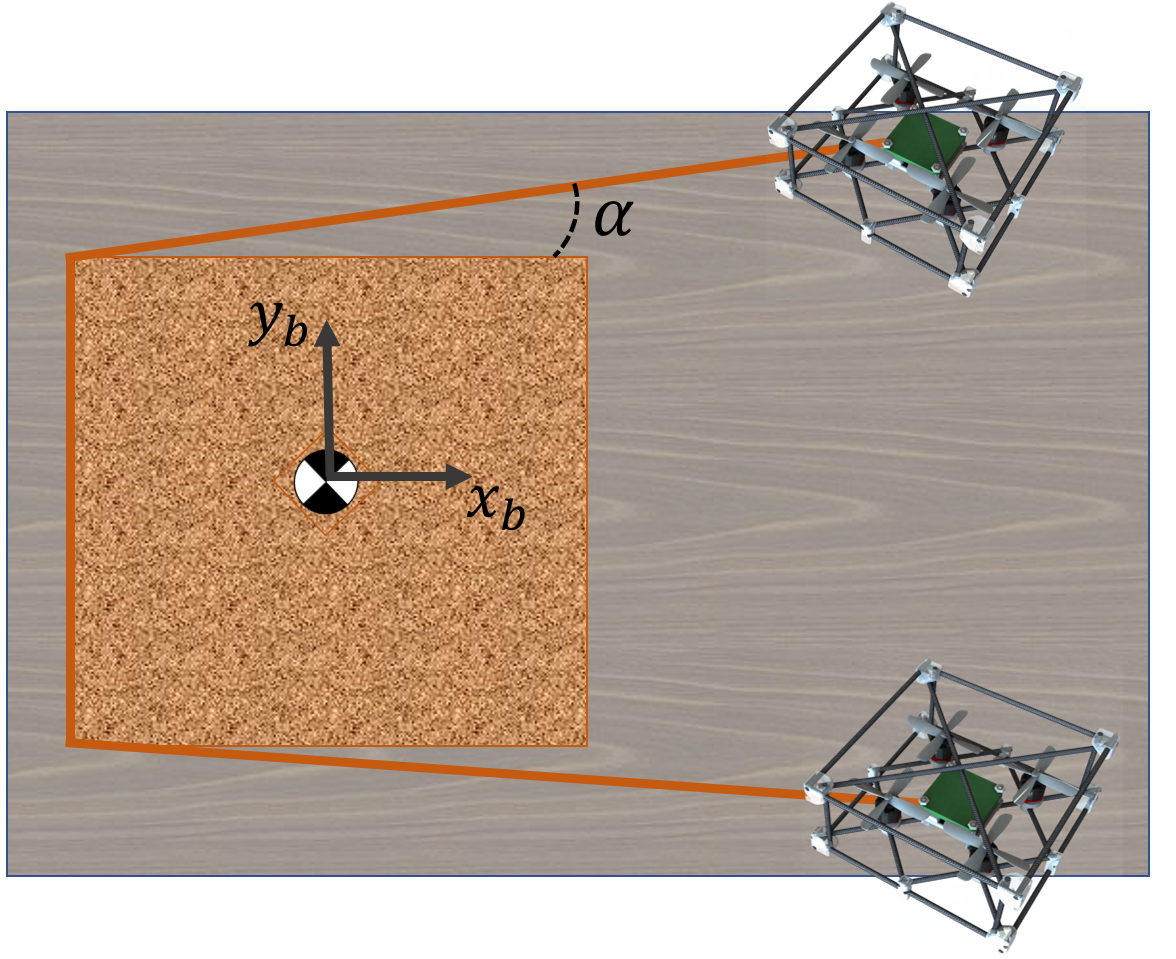}\\
     a) \text{ Catenary robot dragging the box} & b) \text{ Catenary robot rolling the box} & c) \text{ Rolling or dragging top view}
\end{array}
$
\caption{Free body diagram of the catenary robot interacting with the box while (a) dragging. (b) rolling (c) Top view.}
\label{fig:boxDynamic}
\end{figure*}

During this state,
each quadrotor needs to maintain the contact point $\mathbf{p}_i$ and its location with respect to the box
\begin{equation}
{}^{b}\mathbf{r}_1 = {}^{b} \mathbf{p}_1 + \ell_1 \mathbf{Rot}(z,\alpha) \mathbf{Rot}(x,\gamma) \mathbf{e}_1\new{,}
\end{equation}
where $\alpha$ is the angle formed between the box and the cable from the top view as shown in Fig. \ref{fig:boxDynamic}(c), $\gamma$ is the angle on the contact point to the quadrotors shown in Fig. \ref{fig2:contact}, $\ell_1$ is the euclidean distance from the point of contact to the quadrotor, and $\mathbf{Rot}(a,b)$ is a rotation on axis $a$ of an angle~$b$.
Since ${}^{\new{w}}\mathbf{\dot{R}}_{b} = {}^{\new{w}}\mathbf{{R}}_{b} \hat{\boldsymbol{\omega}}_b$, we can compute the 
 the quadrotor's trajectory for position, velocity\new{,} and acceleration in the world frame, as a function of the box trajectory ${}^w\mathbf{r}_{1}^d(t),$
\begin{eqnarray}
            {{}^{w}\mathbf{r}_{1}^d} &=& 
             {}^{w} \mathbf{r}_b^d 
             + {}^{w}\mathbf{{R}}_{b}\;  {}^{b}\mathbf{r}_{1}, \\
            {{}^{w}\mathbf{\dot{r}}_{1}^d} 
            &=& 
            {}^{w}\mathbf{\dot{r}}_b^d
            +  {}^{w}\mathbf{\dot{R}}_{b} \;  {}^{b}\mathbf{r}_{1}, \nonumber\\
            &=& 
            {}^{w}\mathbf{\dot{r}}_b^d
            +  {}^{w}\mathbf{{R}}_{b} \hat{\boldsymbol{\omega}}_b\;  {}^{b}\mathbf{r}_{1}, \\
           {{}^{w}\mathbf{\ddot{r}}}_{1}^d 
           &=& {}^{w}\mathbf{\ddot{r}}_b^d
             + {}^{w} \mathbf{\ddot{R}}_{b}^d   \;  {}^{b}\mathbf{r}_{1}, \nonumber\\
             &=& 
            {}^{w}\mathbf{\ddot{r}}_b^d
             + {}^{w} \mathbf{{R}}_{b}^d ( \hat{\boldsymbol{\omega}}_b^2 +
             \dot {\hat{\boldsymbol{ \omega}}}_{b}
             )\; {}^{b}\mathbf{r}_{1}.
 \end{eqnarray}

The values of $\mathbf{p}_i$ and ${}^{w}\mathbf{\dot{R}}_{b}$ depend on the action\new{,} since $\mathbf{p}_i$  determines the arm vector that generates the torque for rolling, and ${}^{w}\mathbf{\dot{R}}_{b}$  determines the direction of the tension. We expand this discussion in the next section.

It is important to highlight that the configuration space changes depending on the action. During the dragging action, the configuration space is $SE(2)\times SO(3)^2\times (S^{1})^{4}$.
$SE(2)$ is defined by the translation of the box on the $xy$-plane and the yaw angle of the box $\psi$. 
$SO(3)^2$ is defined by the quadrotors' orientation. 
$(S^{1})^{4}$ is defined by the two universal joints formed by the contact of the cable on the box.
During the rolling action, the configuration space is $SE(2)\times SO(3)^2\times (S^{1})^{5}$, note that the only change is given by adding a rotation angle in which the box is going to be rolled.

These three states apply for the two types of actions.
An important factor to consider is the friction on the floor,
if the friction is low, it might be easier to drag, but if it is high, rolling would require less effort for the quadrotors.

Once the cable looses contact with the box, due to either cable slips or transportation task completion, the robot comes back to the \textit{Free-catenary state.}

\section{Considerations and Control Design}
In this section, we explain the relevant considerations to choose between dragging and rolling. Then we show the control design to make the box track a desired trajectory.
 
\subsection{Contact points for dragging}
In order to translate and rotate the box during the dragging action, we need to control $\psi$ and $\mathbf{r}_b$ on the $xy$-plane.
We illustrate the side view of the dragging action in Fig.~\ref{fig:boxDynamic}$(a)$.
In this case, we focus on generating pulling forces $\mathbf{t}_i$ such that they maximize the force in the direction of motion.

To perform dragging, the catenary robot places the points of contact $\mathbf{p}_i$ lower to the height of the box's center of mass, and in the opposite edge of the desired movement direction. This maximizes the component of the force applied in the direction of the movement and minimizes torques in the box as 
$
         {}^{b}\mathbf{p}_1 = \left [\ -l/2, \, -w/2, \, z_{drag} \right ]. 
$   
Therefore, placing the quadrotors in appropriate locations is important.
We use angles $\gamma$ and $\alpha$ that are always referred to the contact point and the quadrotors.
For forward motion, the best height $z_{drag}$ would be as low as possible, i.e., $z_{drag}=-h/2$, with $\alpha=0$ and $\gamma=0$, i.e., close to the ground and the tension pointing forward.
Then, all the tension from the quadrotors can be used to move the box forward. However, the angle $\gamma=0$ cannot be achieved at the minimum point because that would imply that the quadrotor is on the ground. Additionally, the angle $\alpha=0$ does not maximize the torque that controls $\psi$. For small values of $\alpha$, the prop-wash from the quadrotors generate  forces in the opposite direction of motion, increasing the difficulty to manipulate the box.

\subsection{Contact points for rolling}
For rolling the box, torque must be maximized.
Here, the friction force $\mathbf{f}_b$ contributes to the total torque.
Thus, the catenary robot should be placed at the corner to reach the maximum distance between the center of mass and the contact point $\mathbf{p}_i$ which maximizes the moment arm $d_f$ and the force should have an angle $\beta=\pi/2$ with respect to the moment arm $d_f$.
However, to avoid the catenary robot to slip and lose the maneuver we set $z_{roll}$ as the distance bellow the corner and the angle $\beta$ to be almost perpendicular to the arm moment as shown in Fig. \ref{fig:boxDynamic}$(b)$.

Here, $\mathbf{t}_i$ is directly affected by angle $\alpha$ shown in Fig.~\ref{fig:boxDynamic}(c).
This angle should be minimized for the catenary robot to apply the maximum force in the direction of the desired movement.
However, at this point the downwash generates an opposite force over the box and $\alpha=0$ is not longer optimal.
So, it is necessary to find an angle $\alpha$ where $\mathbf{t}_i$ is maximized, but at the same time the downwash effect is minimized.

Vertical slipping between the cable and the box is considered.
Although, there is no dynamics or control law to work on this case.
We assume that whenever slipping occurs, the catenary robot goes back to free mode and starts  the maneuver again.
The contact point for rolling action is ${}^{b}\mathbf{p}_1 = \left [\ -l/2, -w/2, z_{roll} \right ]\new{,}$ and placing of the quadrotors with respect to the box performing translations and rotations are as follows
\begin{equation}
         {}^{b}\mathbf{r}_1 = {}^{b} \mathbf{p}_1 + \ell_1 \new{\mathbf{Rot}(z, \alpha)} \new{\mathbf{Rot}(x, \gamma)} \mathbf{e}_1\new{.}
\end{equation}
Then, quadrotors positions in the world frame are given by
\begin{eqnarray}
            {{}^{w}\mathbf{r}_{1}^d} &=& 
             {}^{w} \mathbf{r}_b^d 
             + {}^{w}\mathbf{{R}}_{b}^d(\psi){}^{w}\mathbf{{R}}_{b}(\phi)
              {}^{b}\mathbf{r}_{1}.
\end{eqnarray}          

\subsection{Control Design}

\begin{figure*}[t]
    \centering
    \includegraphics[height = 2.5cm]{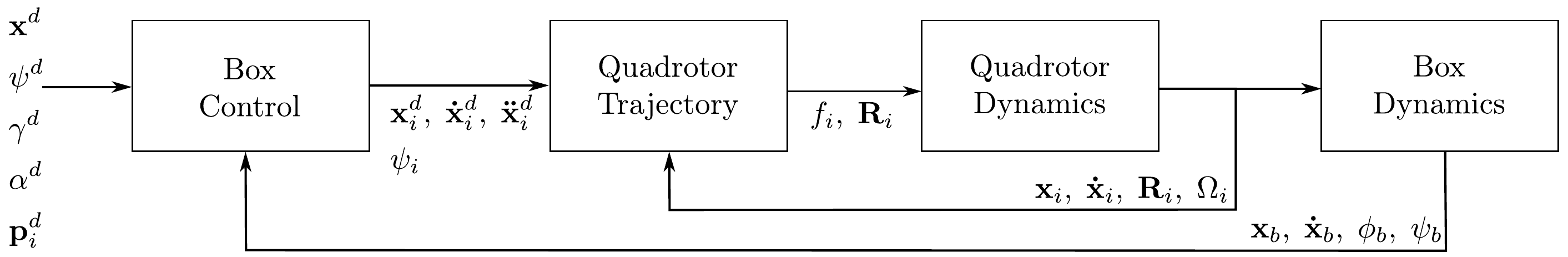}
    \caption{Control architecture for manipulating cuboid objects with catenary robot.}
    \label{fig2:diagramcontrol}
\end{figure*}

To drive the box through a trajectory defined by the position $\mathbf{r}_b^d(t)$, we compute errors in position and velocity of the box as
$\mathbf{e}_p={}^{w}\mathbf{r}_b^d-{}^{w}\mathbf{r}_b$ and $\mathbf{e}_v={}^{w}\mathbf{\dot{r}}_b^d - {}^{w}\mathbf{\dot r}_b$ which are then used for computing a PID controller as 
\begin{equation}
      \mathbf{f}_{b}^{d}=\mathbf{K}_p \mathbf{e}_{p} + \mathbf{K}_\new{v} \mathbf{e}_{\new{v}} + \int{\mathbf{K}_i \mathbf{e}_{i}} + m_b\mathbf{\ddot{r}}_b^d,
\label{eq:Fdes}
\end{equation}
where $\mathbf{K}_p$, $\mathbf{K}_i$\new{,} and $\mathbf{K}_\new{v}$ are appropriate diagonal gain matrices. The integration term here serves as a suppression of the error generated by ignoring the magnitude of the tension applied by the box. This force is then translated as the desired acceleration of each quadrotor in $\{w\}$. The design of this quadrotor trajectory and controller is shown in previous work \cite{catenaryrobot}.
\new{For orientation $\psi^d(t)$ we compute the \textit{yaw} angle of each quadrotor}
\begin{equation}
      \psi_i =  \mathbf{K}_p \mathbf{e}_{\psi} + \mathbf{K}_\new{v} \mathbf{e}_{\dot{\psi}}\new{.} 
\label{eq:Fdes}
\end{equation}
Then, desired rotation is defined by computing the desired angle of the box $\psi^d$ as
\begin{equation}
    {}^{w}\mathbf{{R}}_{b}^{d} = \mathbf{{Rot}}(z,\psi^d)\new{.}
\end{equation}
%
% \section{tmp}
From (3) and (4), the tension can be isolated to get the following
\begin{align}
  \mathbf{t}_{1} &= {}^{{w}}\mathbf{R}_B^\top(- m_{b} {\mathbf{\ddot{r}}}_{b} -m_{b} g \mathbf{e}_3  +{}^{{w}}\mathbf{R}_{B}\mathbf{f}_b)\new{,}\\ 
\mathbf{t}_{1} &=\hat{\mathbf{p}}_{1}(
%\mathbf{f}_{b}\times \mathbf{d}_b
\mathbf{J}_b{\boldsymbol{\omega}}_b \times \boldsymbol{\omega}_b - 
\mathbf{J}_b\dot{\boldsymbol{\omega} }_{b})\new{.}
\end{align}
Then, we can use the geometric controller in \cite{lee2010geometric} to generate the tension and track the trajectory from (6)-(8).

%-------------------------------------------------------------
\section{Experiment Results}
We designed three different experiments to test our method. First, we test the dragging action. We translate and rotate the box on the surface. Second, we test the rolling action. The catenary robot pulls the box to rotate it around the $y$-axis. Third, we test switching between dragging and rolling. We drag the box to the desired distance and then switch to roll action.

{Experimental Testbed\footnote{The source code for simulation and implementation is available at\\ \url{https://github.com/swarmslab/Catenary\textunderscore Robot}}: }
We use a custom Crazyflie quadrotor based on the Crazybolt board with brushless motors for a higher payload. We use a motion capture system (Optitrack) at 120 Hz to localize the quadrotors and box. The internal IMU sensor gives the information of the angular velocities. We use a cable of length $\ell=1$ m, a box with dimensions $w=l=h= 0.155$ m and weight of $80$ g.

\begin{figure}[t]
    \centering
    {\includegraphics[width=0.4\textwidth]{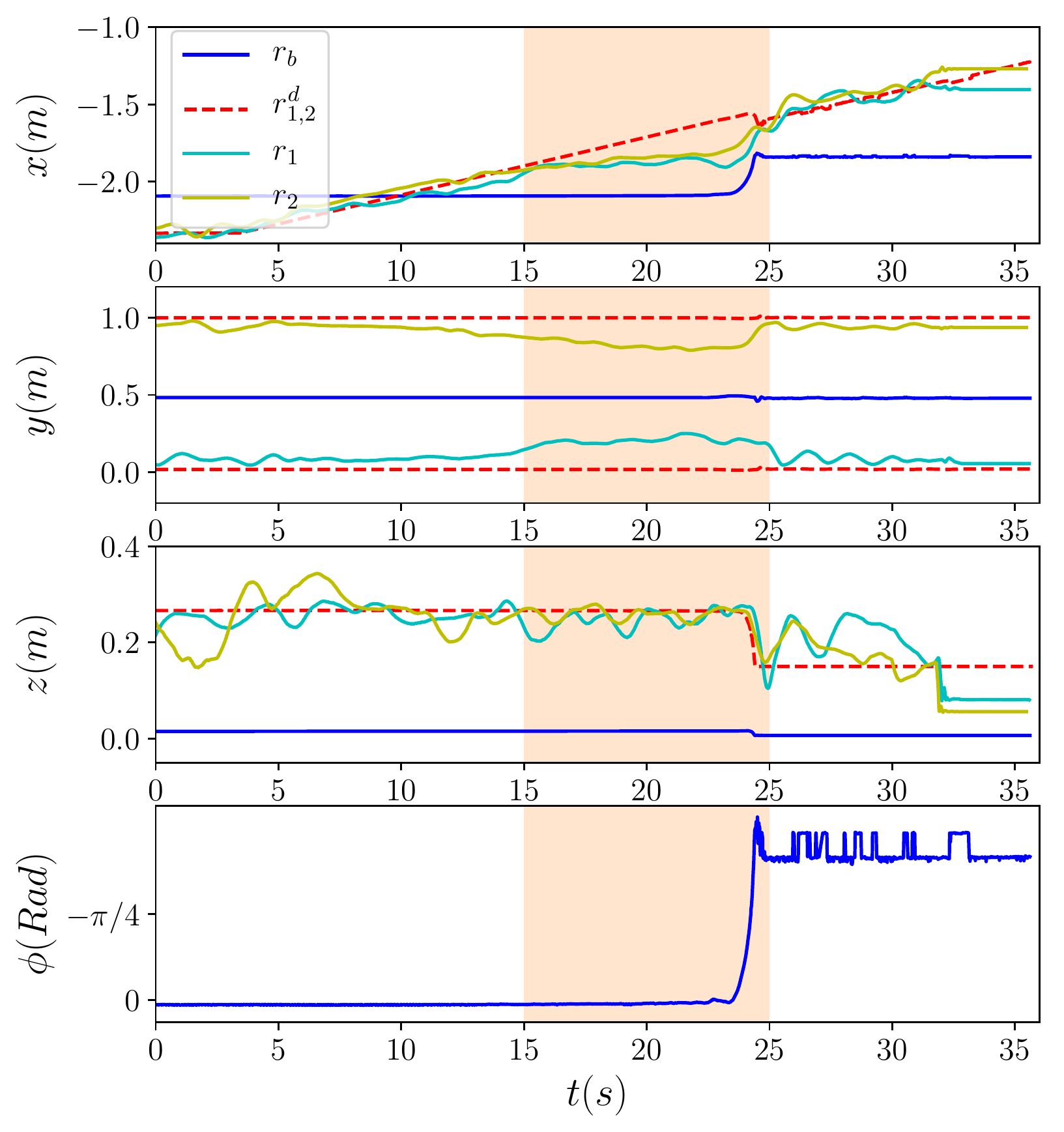}
    \caption{Rolling experiments result: Rolling a box with the catenary robot. Position of the quadrotors and the box in solid lines while desired trajectories are in dashed red lines.}
    \label{fig:rolling_trajectory}}
\end{figure}

\paragraph{Rolling experiment}
As we described previously, the contact point has to be as high as possible.
We choose $z_{roll}=12$ cm to leave a small gap of $3.5$ cm between the contact point and the top of the box, in case of small slips of the cable. 
On one side, when the $\beta$ angle is close to $\pi/2$, it maximizes the torque, but it is easy to slip and lose contact.
Satisfying that $\beta$, the angle $\gamma$ would be $\gamma=\beta-\arctan(12.0/15.5)=0.91$.
On the other side, a small angle $\beta$ minimizes the slipping when $\gamma$ is close to zero, but the torque is reduced, requiring more effort from the quadrotors.
We tested multiple $\gamma$ angles, finding that angles below $\pi/6$ reduce the slipping.
In our final experiment, we defined $\gamma=\pi/12$, because this was the largest reliable angle that reduces the slipping during the rolling action.
The angle $\alpha$ is $\pi/4$, because it reduces the force of the prop-wash in the opposite direction of the motion.
%We arbitrarily , desired angles $\alpha^d= \pi/4$, $\gamma^d= \pi/12$, and $\beta^d= \gamma^d + \gamma_b$, where $\gamma_b$ is the actual tilting angle of the box.
In Fig.~\ref{fig:rolling_trajectory}, it is shown how the catenary robot makes the box track the trajectory $x_{b}^{d}(t)=t$, $y_{b}^{d}(t)=0.5$, and drives $\phi_{b}^{d}$ from $0$ to $\pi/2$. The vertical orange gap indicates the time interval when the cable is pulling the box to perform the rolling action.
Errors in \new{$x$-axis} and \new{$y$-axis} get bigger when the cable starts to apply tension on the box.
As the required tension is unknown the controller is not able to maintain error close to zero.
It is possible to see that in the \new{$z$-axis} the controller drives the quadrotors downwards until it reaches a predefined minimum height to avoid hitting the floor set at $0.15\, m$.
% goes forward and backwards making the box to roll in the pitch angle from $0\rightarrow \pi/2$, then $\pi/2 \rightarrow 0$ twice. It is noticeable a greater error in the $z$-\textit{axis} due to the release of the tension creates a sudden acceleration.

\begin{figure}[t]
    \centering
    {$
    \begin{array}{c}
         \includegraphics[width=0.4\textwidth]{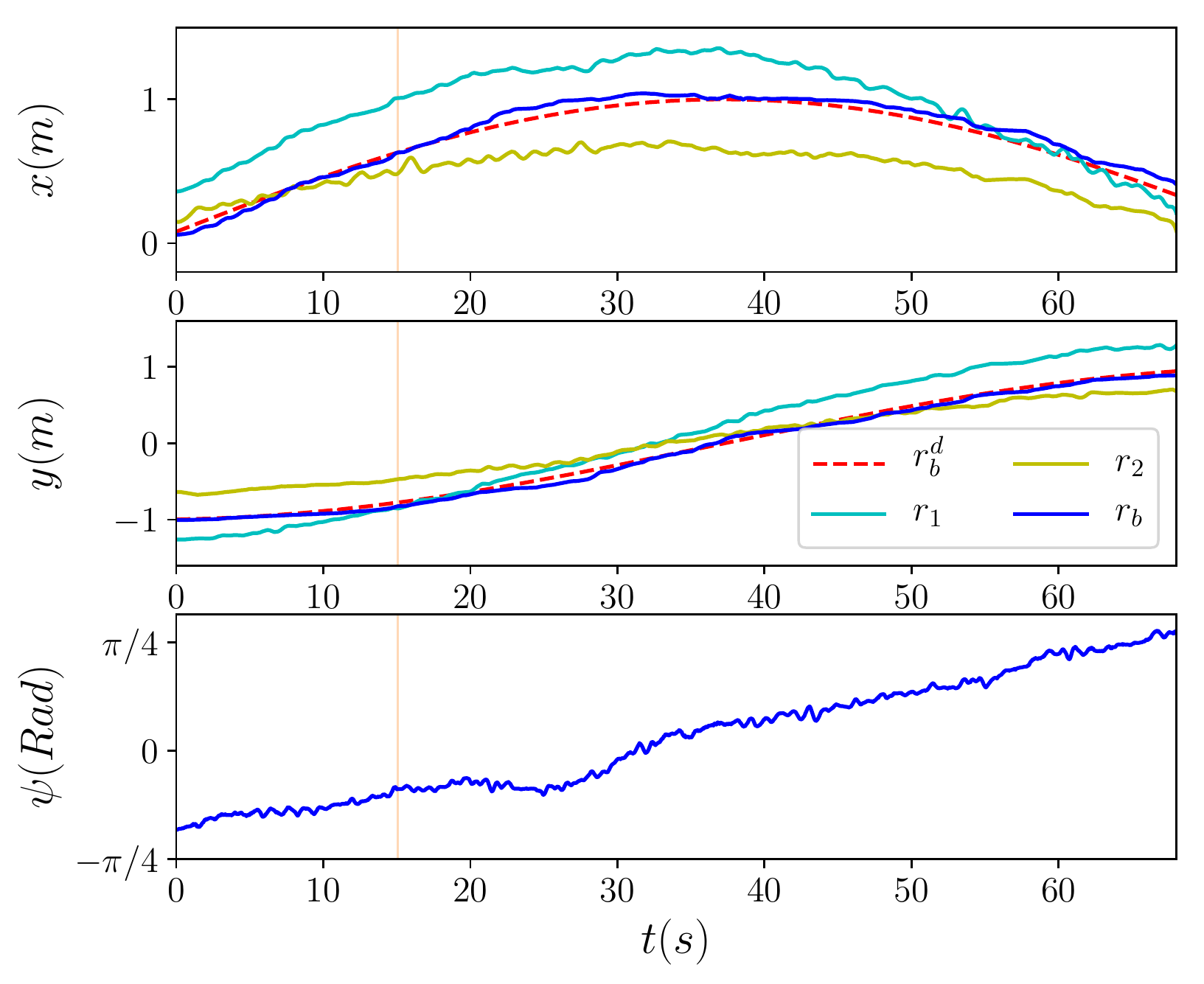}  \\
         \includegraphics[width=0.4\textwidth]{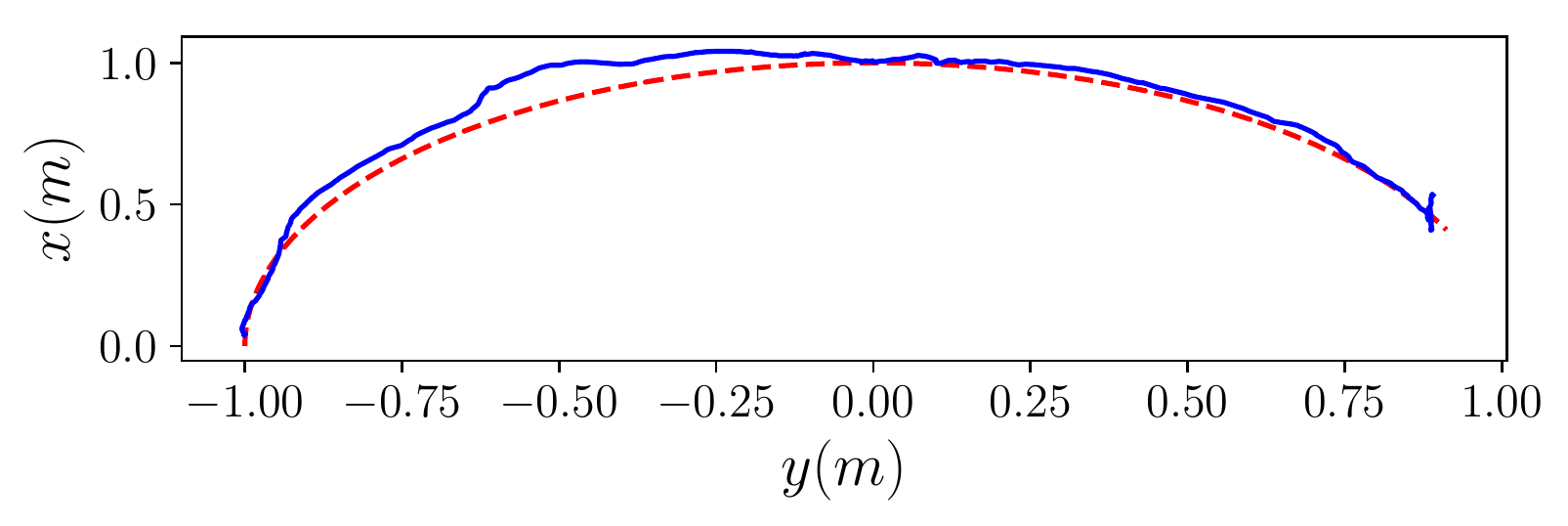}
    \end{array}
    $
    \caption{Dragging experiment result: Dragging a box with the catenary robot in a semicircular trajectory. Position of the quadrotors and the box in solid lines while desired trajectories are in dashed red lines. The bottom figure shows the desired trajectory and the position in the $xy$-plane}
    \label{fig:dragging_trajectory}}
\end{figure}

\paragraph{
Dragging experiment}
In dragging action, we found that the contact point between the cable and the box should be close to the box base for the catenary robot to reduce the potential rolling. The desired angles of the contact point are $\gamma^d= \pi/12$, and $\alpha^d=\pi/4$. Note that $\alpha$ is not zero due to the prop-wash generating a force in the opposite direction on the box. 
In Fig.~\ref{fig:dragging_trajectory}, the desired trajectory for the box is set as $x_{b}^{d}(t)=\sin(t)$, $y_{b}^{d}(t)=\cos(t)$, $z(t)=0$, and $\psi^d(t)= \arctan(y,x)$.
The position of the box tracks closely the semicircular trajectory.
It can be seen that quadrotors modify its trajectory to correct box position and orientation. The average error in position is $\mu_x= 52e-3$ and $\mu_y=0.2425$, and its standard deviations $\sigma_x=0.30$ and $\sigma_y=0.47$, .
The error for yaw-angle is $\psi(t)$  $\mu_{\psi}= 12e-2$, and their standard deviation $\sigma_{\psi}= 6e-3$. 

\begin{figure}[t]
    \centering
    {\includegraphics[width=0.4\textwidth]{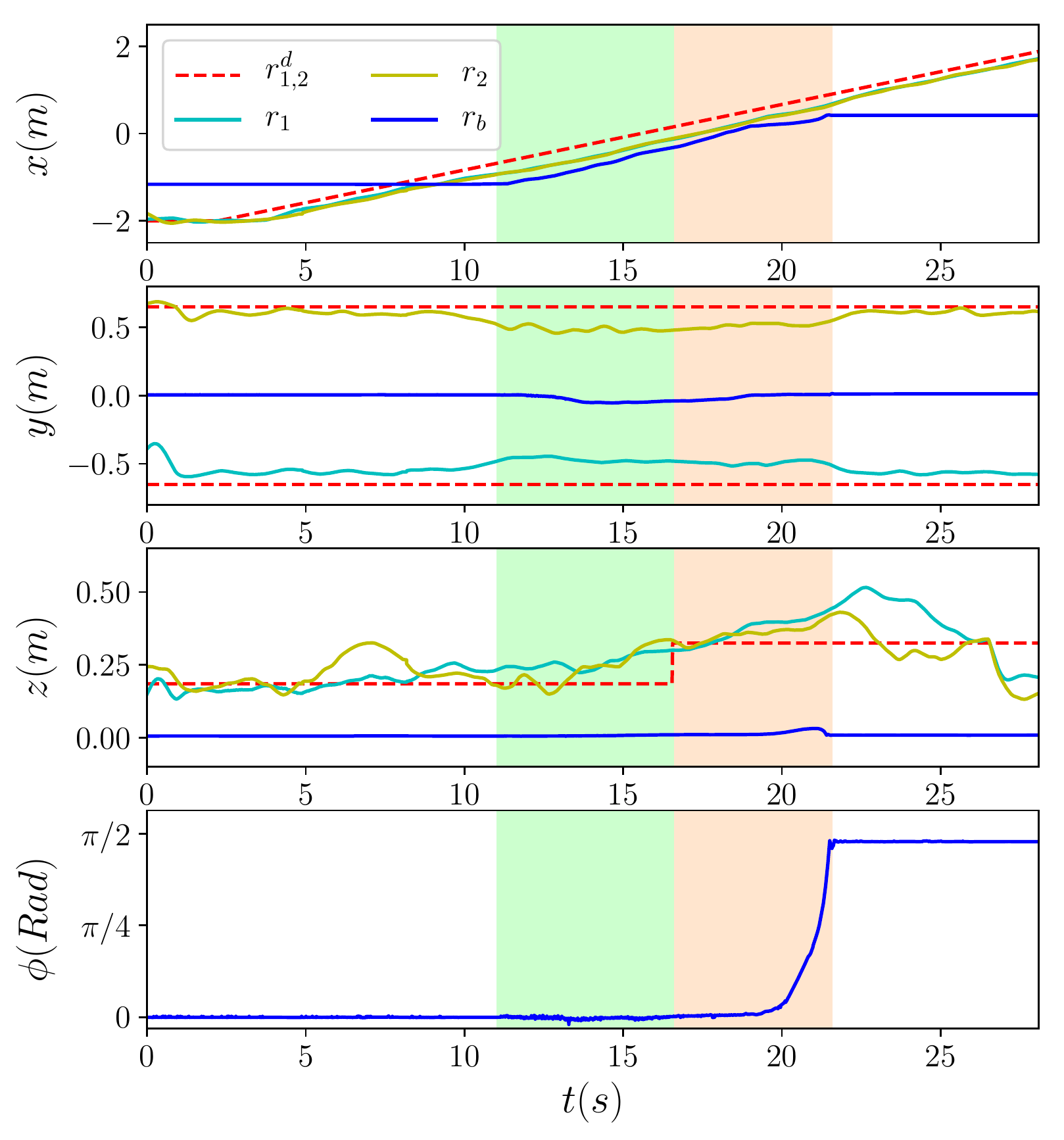}
    \caption{Dragging and then rolling experiment result: Dragging and then rolling a box with the catenary robot. Position of the quadrotors and the box in solid lines while desired trajectories are in dashed red lines.}
    \label{fig:rolling_dragging_trajectory}}
\end{figure}

\paragraph{
Dragging and then rolling experiment}
For this experiment, we start \textit{moving free} from 0 to 11s, then we switch to \textit{dragging} from 11s to 16.5s, and, finally, switch again to \textit{rolling}.
In Fig.~\ref{fig:rolling_dragging_trajectory}, all of the stages are shown and marked by color regions.
The greatest errors in $xy$-plane are when the catenary robot is \textit{dragging},
while the greatest error in $z$-axis is when the catenary robot is \textit{rolling}. Which are caused by lack of feedback about the tension experienced by the quadrotors caused by the box.
%-------------------------------------------------------------
\section{Conclusions and Future Work}
In this work, we use a catenary robot to manipulate a box on a planar surface
based on control strategies that take advantage of the natural shape of
hanging cables due to their weight, i.e., catenary.
We design two strategies for two manipulation modes of boxes, dragging and rolling. When friction force with the ground is lower, dragging is advantageous. In contrast, performing rolling is difficult when friction forces do not generate enough torque.
However, if friction constraints are met, we design a controller able to
place the cable in different contact points and perform either dragging or rolling arbitrarily.
The strategy depends on the contact point and angles between the box and the quadrotors.
Three experiments were implemented to show the performance of the controllers manipulating an actual box.
In future work, we will consider the interaction of multiple catenary robots to manipulate the box on the ground and lift it and manipulate it in the air.
%-------------------------------------------------------------
\bibliographystyle{ieeetr}
\bibliography{referencias.bib}

\end{document}